\documentclass{ieeeaccess}
\usepackage{cite}
\usepackage{amsmath,amssymb,amsfonts}
\usepackage{textcomp}
\usepackage{multirow,multicol}
\usepackage{algorithm,algpseudocode}
\usepackage{graphicx}
\usepackage{color}
\usepackage{booktabs}
\usepackage{subfig}

\def\BibTeX{{\rm B\kern-.05em{\sc i\kern-.025em b}\kern-.08em
    T\kern-.1667em\lower.7ex\hbox{E}\kern-.125emX}}
\begin{document}
\history{Date of publication xxxx 00, 0000, date of current version xxxx 00, 0000.}
\doi{10.1109/ACCESS.2022.0092316}

\title{Geometric Pooling: maintaining more representative information}
\author{\uppercase{HAO XU}\authorrefmark{1}, \uppercase{JIA LIU}\authorrefmark{1}, 
\uppercase{YANG SHEN}\authorrefmark{1}, \uppercase{KENAN LOU}\authorrefmark{1}, 
\uppercase{YANXIA BAO}\authorrefmark{1}, \uppercase{RUIHUA ZHANG}\authorrefmark{1}, 
\uppercase{SHUYUE ZHOU}\authorrefmark{1}, \uppercase{HONGSEN  ZHAO}\authorrefmark{1},
%\uppercase{ZHEN YE}\authorrefmark{1},
\uppercase{XINMIAO  ZHU}\authorrefmark{1},
\uppercase{SHUAI WANG}\authorrefmark{2}}

\address[1]{Lishui University, Lishui 323000, China}
%\address[2]{Zhejiang Xiaomache Technology Co.,Ltd., Lishui 323000, China}
\address[2]{CNRS, LPL, Aix Marseille University, Aix-en-Provence, 13100, France}
\tfootnote{This work was supported in part by the Zhejiang Provincial Natural Science Foundation of China under Grant LY21F020004, in part by the National Natural Science Foundation of China Grant No. 61572243,  in part by the Public Welfare Technology Application Research Program Project of Lishui City under Grant No.2022GYX12 and 2022GYX09, in part by the Intelligent Manufacturing Technology Innovation Base Project of Lishui City under Grant HXZKA2022010 and  Construction Project of Lishui University Discipline (Zhejiang Province First Class Discipline, Discipline name:intelligent science and technology) No.XK0430403005.
%in part by the Baishanzu National Park Scientific Research Program under Grant No.2021KFLY07.
}

\markboth
{Author \headeretal: Preparation of Papers for IEEE TRANSACTIONS and JOURNALS}
{Author \headeretal: Preparation of Papers for IEEE TRANSACTIONS and JOURNALS}

\corresp{Corresponding author: Kenan Lou (nanekuol@126.com).}

\begin{abstract}
Graph Pooling technology plays an important role 
in graph node classification tasks. 
Sorting pooling technologies maintain large-value units for 
pooling graphs of varying sizes.
However, by analyzing the statistical characteristic
of activated units after pooling, 
we found that a large number of units dropped by sorting pooling 
are negative-value units that contain representative information and 
can contribute considerably to the final decision.
To maintain more representative information, 
we proposed a novel pooling technology, 
called Geometric Pooling (GP), 
containing the unique node features with negative 
values by measuring the similarity of all node features. 
We reveal the effectiveness of GP from the entropy reduction view.
The experiments were conducted on TUdatasets to show the effectiveness of GP. 
The results showed that the proposed GP 
outperforms the SOTA graph pooling 
technologies by $1\%$ $\sim$ $5\%$ with fewer parameters.
\end{abstract}

\titlepgskip=-21pt
\maketitle

\section{Introduction}
The graph neural networks (GNNs) \cite{Bronstein2017GDL,Scarselli2009GNNM} 
have been applied and developed
in many domains, such as Biomedicine \cite{Duvenaud2015CNGLMF,Glimer2017MMPNNs}, 
Semantic Segmentation \cite{Landrieu2018SPG}, Pose Estimation \cite{xiao2020spcnet, xiao2022adaptivepose, xiao2022adaptivepose++, xiao2022learning, xiao2022querypose},
and Recommender System \cite{Ying2018pinsage}.
GNNs can extract good representations 
from graph-structured data in non-Euclidean space.
Mimicking Convolutional Neural Networks (CNNs), 
researchers designed GNNs sequentially containing convolutional filters, 
activation functions, and pooling operators to 
extract features with non-linearity and invariance.
Nevertheless, the typical pooling technology in CNNs, such as max-pooling and average-pooling,  requires that the processed data are structured (i.e., the data in a neighborhood are in a strong correlation assumption) and is not suitable for non-Euclidean data. 
However, the graph-structured data does not satisfy the assumption,
and the pooling technology brings the key invariance and 
increases the receptive field of the whole graph, 
is very important for GNNs.

Many researchers paid attention to the GNN pooling technologies, 
which are divided into Global Pooling \cite{Zhang2018DGCNN,Vinyals2016Sets2Sets}, 
Topology-based Pooling \cite{Inderjit2007WGCwoE,Defferrard2016FLSF}, 
and Hierarchical Pooling \cite{Ying2018DiffPooling,gao2019graphUnet}.
Global pooling is a simple and effective technology, 
which is a feature-based technology differing from the topology-based pooling technology.
Global pooling methods use manual summation or learning methods to pool all the representations
of nodes in each layer. Although global pooling technology does not explicitly consider 
the topology of the input graph, the previous graph convolutional filter has already 
considered the topology and implicitly influenced the later pooling.
Hence, this paper focuses on the global pooling methods because of 
its computational efficiency and universality.

\begin{figure*}
	\centering
	\includegraphics[scale=0.4]{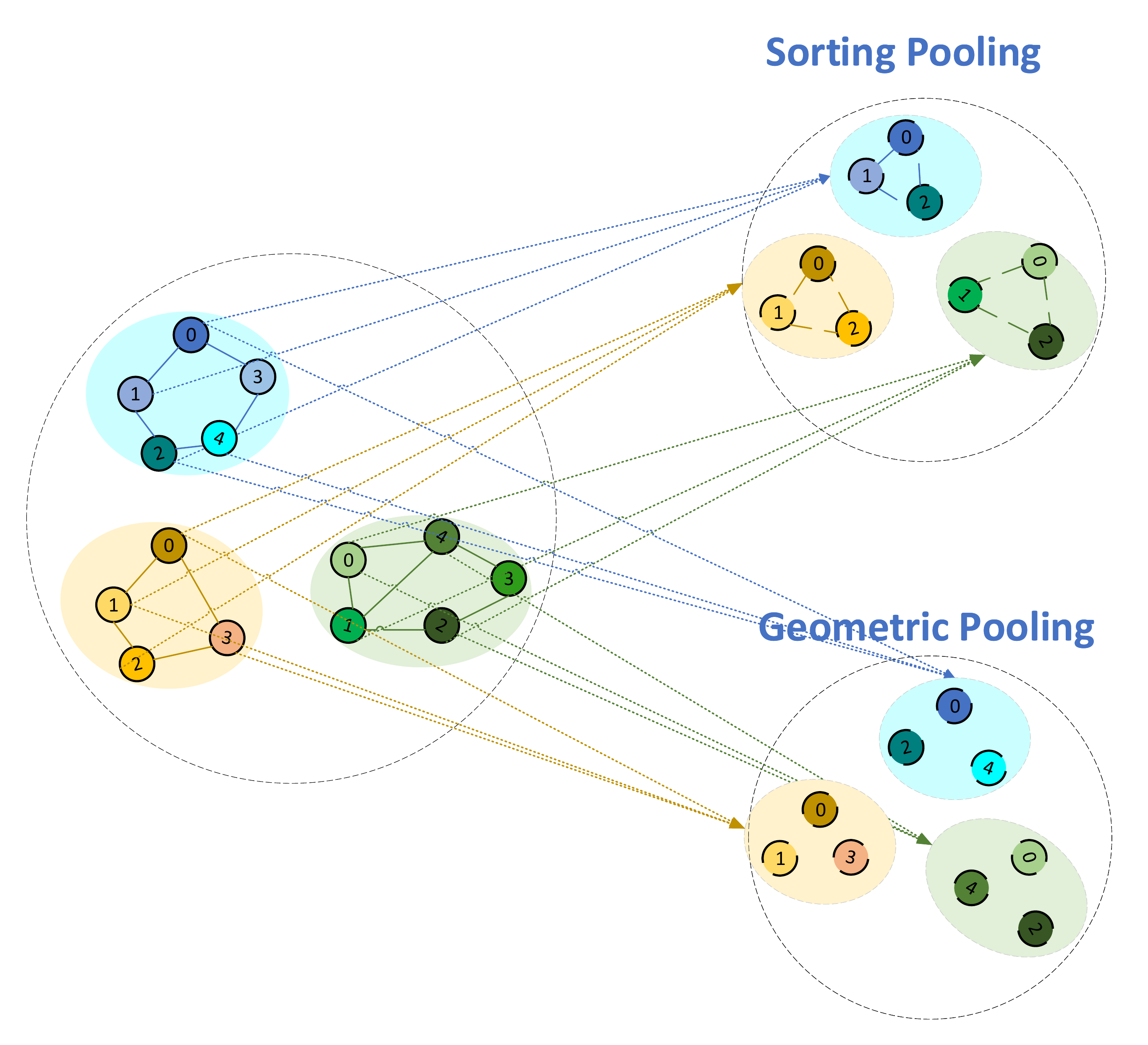}
	\caption{We show the main difference between sorting pooling and the proposed GP.
		Different graphs containing different-number nodes are marked in different colors.
		The numbers in different nodes represent the {\color{black}amplitude-sorting} indices.
		For example, the amplitude of the node indexed by 0 is larger than that of indexed by 1.
		Meanwhile, the color contrast among different nodes in a particular graph means 
		the similarity among different nodes.
	  {\color{black}If three nodes are retained, 
        Sorting Pooling keeps the three nodes with indices 0, 1, and 2, 
        while GP   retains the three nodes with the lowest similarity.}}
	\label{fig:highlight}
\end{figure*}

However, the existing global pooling methods dropped much useful information due to 
their decision criteria and the activated function, such as \textit{tanh}, used in the GNNs.
For example, Sorting Pooling \cite{Zhang2018DGCNN} 
maintains the large-value units and drops small-value units 
by sorting the units. The effectiveness of Sorting Pooling 
is depended on the assumption that
the dropped units contribute little to
the final decision.
However, due to the activation function, \textit{tanh},
used in Sorting Pooling, 
the sorting methods not only drop the 
small-value units but also the negative-value ones.
The negative-value units with large absolute values
actually carry considerable information, 
representing the node and contributing much to the final decision.
And unfortunately, \textit{tanh} cannot be replaced by the positive-value activation functions 
(such as ReLU, ReLU6) because this leads to a performance drop.
\footnote{{\color{black}
To verify this point, we replaced the \textit{tanh} activation function with ReLU in DGCNNs, 
and results showed a performance drop. 
We have given the experiment result in Appendix \ref{appendix:sec:activation}
}}

To explore the number of dropped negative-value units and 
the small-value units, 
which respectively contribute much and little to the final decision, 
we counted and analyzed their number in Sec \ref{subsec:analysis}.
Obviously, the ``useless'' units accounted for 
a small proportion of all dropped units,
and ``useful'' units accounted for a larger proportion. 
The dropped information problem cannot be 
addressed by a simple absolute sorting pooling because 
when the number of dropped units is fixed (Seeing \cite{Zhang2018DGCNN}), 
the number of ``useless'' units is not enough and the useful units are also dropped.

{\color{black}
Measuring the contribution of a unit is key to improving graph pooling. Motivated by the pruning technologies, which also study the measurement of 
units and kernels, we explored a novel measurement FPGM \cite{2019CVPRFilterPruning} and }
proposed a novel pooling decision criterion - 
Geometric Pooling (GP) to maintain more representative information, 
whether positive-value or negative-value units. 
Different from the global pooling methods, which drop the units with relatively 
less contribution, the proposed GP dropped the most replaceable units.
Various metrics, such as Euclidean distance, inner product, cosine similarity \cite{Tan2018datamining}, and Mahalanobis distance\cite{Mahalanobis1999distance} can be used to measure the similarity between different units. Euclidean distance was selected in the present study due to its simplicity and low computation cost.
In the graph classification task, we dropped the most replaceable node features.
The main difference between sorting pooling and GP is visualized in Figure \ref{fig:highlight}.
By exploring the effectiveness, we found that
the proposed GP can be viewed as a sort of regularization method, entropy reduction, 
such as Label Smoothing \cite{Mller2019WhenDL}, Mixup\cite{zhang2018mixup}, 
and Cutmix \cite{yun2019cutmix}.
They alleviate the over-confidence problem, 
which leads to the over-fitting problem.
Dropping the replaceable units maximizes the 
compressing information rate and improves the representation of the node.
We stated the regularization view in Section \ref{subsec:entropy_reduction}.

To verify the effectiveness of GP, 
we used the backbone of DGCNNs \cite{Zhang2018DGCNN}, and used 
GP to replace the original pooling technology.
The evaluated experiments are conducted on the TUdataset \cite{Morris2020TUdataset}.
{\color{black}
Additionally, we fused the advantages of Sorting Pooling and 
Geometric Pooling technologies (named GP-mixed), 
getting a better evaluation performance 
(detailed in Table \ref{tab:results_deep} and Table \ref{tab:results_kernels}).
GP-mixed showed that GP helps the existing graph pooling methods.
}
The experimental results show that GP outperforms the SOTA graph pooling methods.

The main contributions of this paper are as follows.
\begin{enumerate}
	\item We analyzed the units dropped by the existing global pooling technologies and 
    found that many negative-value units with useful information are dropped.
    Seeing this, this paper proposed a novel pooling decision criterion - 
	Geometric Pooling (GP) to drop the most replaceable units, maintaining more representative information.
	
	\item By exploring the effectiveness, this paper viewed the 
    proposed GP as a regularization method, which smooths 
    the distribution of the outputs and maximizes the compressing information rate,
    addressing the over-fitting problem.
	
	\item Comprehensive evaluated experiments show that GP outperforms 
    the existing deep and kernel-based pooing technologies.
\end{enumerate}

\section{Related Works}
\textbf{Graph global pooling technologies.}
Sorting Pooling \cite{Zhang2018DGCNN} maintains the large-value units and drops the 
small-value units. This is similar to the norm-based pruning technologies, 
which determine the importance of the units by their contributions 
to the final decision. Furthermore, SAGPooling \cite{Lee2019SAGPooling} uses 
self-attention \cite{Ashish2017SelfAttention} to mask the features pooled 
by Sorting Pooling. Different from the Sorting Pooling, 
gPooling \cite{Gao2019GUNets} not only uses the sorting indices to select 
the node features but also the corresponding adjacency matrix.
According to the selected adjacency matrix, the topological links
between the selected nodes are strengthened,
and the unimportant links are dropped.

\textbf{Entropy reduction.}
Entropy reduction appears in many regularization methods.
The typical one is label smoothing \cite{Mller2019WhenDL}.
Label smoothing improves the robustness of the trained network by using 
soft targets that are a weighted average of the 
hard targets and the uniform distribution over labels. 
This is equivalent to the effectiveness of the method which 
encourages the output distribution away from the distribution of the 
hard targets. Following label smoothing, 
many researchers proposed many different modified versions, 
such as Mixup \cite{zhang2018mixup}, Manifold Mixup \cite{verma2019manifoldmixup}, 
CutMix \cite{yun2019cutmix}, 
to strengthen the entropy reduction according to different tasks.

Zhang \textit{et. al} proposed Mixup \cite{zhang2018mixup} that used the linear combination of 
different samples of different classes to augment the training samples,
and label smoothing is also used to assign the label to the augmented training sample.
Manifold Mixup \cite{verma2019manifoldmixup} is modified by Mixup, which randomly augments
the feature by the linear combination of different features.
CutMix \cite{yun2019cutmix} is a regularization method that fuses the advantages of 
Cutout and Mixup.

The above methods encourage the output distribution far away from the distribution of 
the hard targets. This alleviates the over-confidence problem of the 
model output, which results in the over-fitting problem.
The proposed GP is based on an argument that 
the similar features work repetitively for the final decision, 
and we can drop some of them.
This also alleviates the over-confidence problem.

\section{Methods}
\label{sec:methods}
In this section, we first set up a typical GNNs model.
Then the number distribution of the activated units is analyzed in the Section \ref{subsec:analysis}.
According to the conclusion of the analysis, we stated 
the proposed Geometric Pooling (GP) in the Section \ref{subsec:gp}.
Finally, we give the effectiveness of GP from the entropy reduction 
view in Section \ref{subsec:entropy_reduction}.

\subsection{Set Up}
A graph neural network consists of four parts: 1) \textit{graph convolution layers} 
extract nodes' local substructure features and define a consistent node ordering;
2) a pooling technology is used to unify the size of nodes and 
impose invariance on the network; 3) an activation layer, such as \textit{tanh}, is used to give 
the non-linearity to the previous layers; 4) {\color{black}
One or several linear layers are used to process the features extracted by the 
stacked graph modules to finish the corresponding tasks.
}

$\mathbf{X} = \{\mathbf{x}_i\}_{i = 0}^N$ denotes the input set 
and $\mathbf{x}_i \in \mathcal{R}^{n \times d}$, i.e., 
the input $\mathbf{x}_i$ consists of $n$ nodes and each node is a \textit{d}-dimensional vector.
$\mathbf{Y} = \{\mathbf{y}_i\}_{i = 0}^N$ denotes the predict set.
In the graph classification task, $\mathbf{A} \in \mathcal{R}$ is the adjacent matrix.
$\mathbf{A}$ is a symmetric 0/1 matrix, 
and the graph has no self-loops.

\textbf{Graph convolutional layers.} 
Graph convolutional layers aim at learning a mapping 
$\mathit{f}: (\mathbf{H}^{(l)}, \widetilde{\mathbf{A}}) \rightarrow \mathbf{H}^{(l+1)}$, 
where $\mathbf{H}^{(l)}$ is the activation vector in the $l_{th}$ layer and 
$\mathbf{H}^{(0)} = \mathbf{X}$. 
For example, we formulated the graph convolution layers defined in \cite{Zhang2018DGCNN}:
\begin{equation}
\label{eq:graph_convolution}
\mathbf{H}^{(l+1)} = \mathit{\sigma}(\widetilde{\mathbf{D}}^{-1}
\widetilde{\mathbf{A}}\mathbf{H}^{(l)}\mathbf{W}^{(l)}),
\end{equation}
where $\widetilde{\mathbf{A}} = \mathbf{A} + \mathbf{I}$ is 
the adjacency matrix with added self-loops, 
$\widetilde{\mathbf{D}}$ is its diagonal degree matrix with 
$\widetilde{\mathbf{D}}_{ii} = \sum_j \widetilde{\mathbf{A}}$, 
$\mathbf{W}^{(l)} \in \mathcal{R}^{d_l \times d_{l+1}}$ is the 
trainable graph convolution parameters of the $l_{th}$ layers, 
and $\mathit{\sigma}$ is a nonlinear activation function.
In Eq. \ref{eq:graph_convolution}, $\mathbf{H}^{(l)}\mathbf{W}^{(l)}$ is a linear transformation
for the input $\mathbf{H}^{(l)}$ to extract the key information. 
Then $\widetilde{\mathbf{A}}\mathbf{H}^{(l)}\mathbf{W}^{(l)}$ used self-looped adjacency matrix 
to propagate node information to neighboring nodes along with the topological structure. 
Finally, $\widetilde{\mathbf{D}}^{-1}$ is used to keep a fixed feature scale after graph convolution.

\textbf{Pooling.} The Global pooling operator of GNNs can be viewed as a selected operator, 
which selects the representative node features and imposes invariance on 
the model. The global pooling operator considers the importance of each node feature.
{\color{black}Additionally, pooling technology ensures graphs represented by different-number nodes share
the same dimension, which is friendly to the computation of the neural network in 
a parameter-shared manner. Usually, $k$ is used to denote the number of 
the maintained nodes.}

\textbf{Activation function.} A typical activation function is the 
hyperbolic tangent function (\textit{tanh}):
$$z = \frac{e^x - e^{-x}}{e^x + e^{-x}}.$$

\subsection{Analysis of the Number Distribution}
\label{subsec:analysis}
We analyzed two sorts of sorting pooling: 
DGCNNs \cite{Zhang2018DGCNN} and Filter Norm Sorting \cite{Lee2019SAGPooling}.
DGCNNs concatenate the features of different graph convolutional layers
to represent the whole graph. 
In DGCNNs, the contributions of the units are determined by the values of 
the units in the last graph convolutional layer, i.e., 
DGCNN drops the node features with small-value units in the last graph convolutional layer.
However, the contribution of a particular node ought to be jointly
determined by all features of this node, not only the units in the last layer.
This dropped a lot of representative node features (representative information).

\begin{figure}[htb]
	\centering
	\subfloat[sorting pooling]{
		\includegraphics[width=0.4\textwidth]{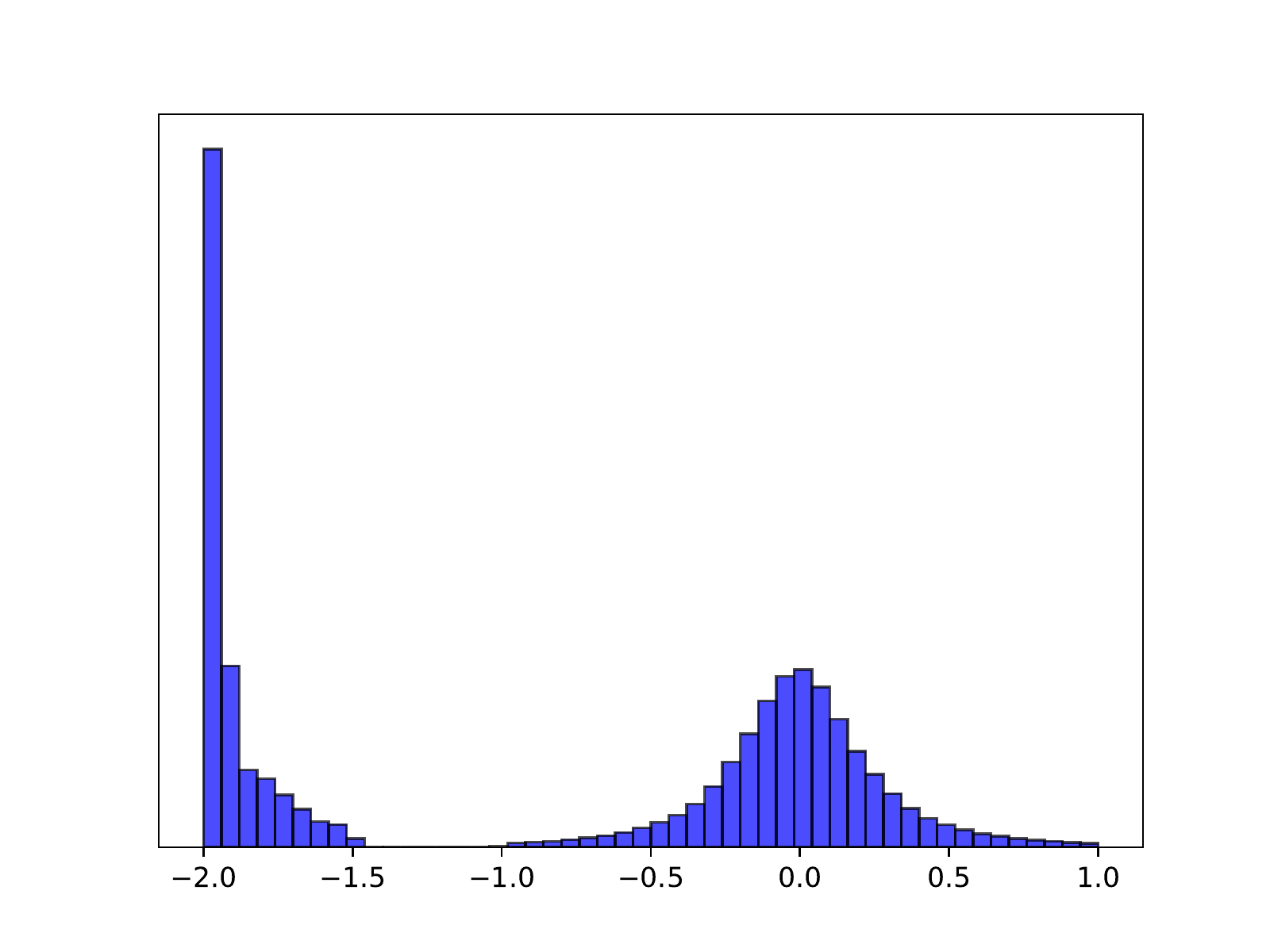}
    \label{fig:subfig:sorting_pooling}}
	\hfill
	\subfloat[geometric pooling]{
		\includegraphics[width=0.4\textwidth]{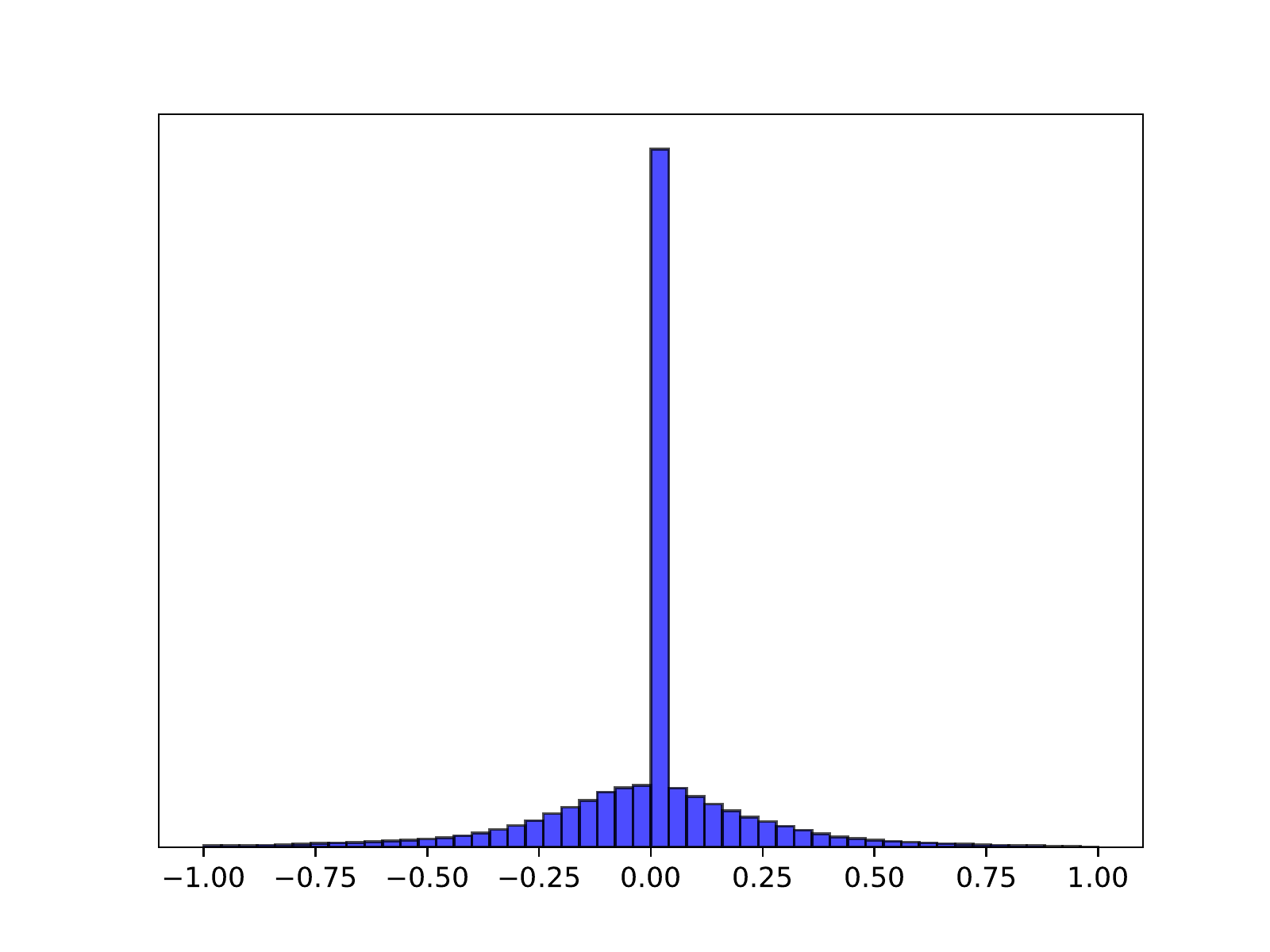}
    \label{fig:subfig:geometric_pooling}}
	\caption{The counting histograms of sorting pooling and GP technologies.
		In \textbf{(a)}, the units are dropped by sorting pooling; 
		in \textbf{(b)}, the units are dropped by geometric distance between the node features.}
	\label{fig:distribution_dropped}
\end{figure}

We counted the values of dropped units in Figure \ref{fig:distribution_dropped}.
We counted the dropped unit on the D\&D dataset (detailed in Section \ref{subsec:exp_set_up}), 
setting $k$ to 20, {\color{black}where k is the number of maintained nodes.} 
As shown in Figure \ref{fig:distribution_dropped} \textbf{(a)}, sorting pooling drops
a large number of negative-value units but with large absolute values.
Differently, the proposed GP drops the node features sharing large 
similarity (Euclidean distance) with other node features.
Obviously, in Figure \ref{fig:distribution_dropped} \textbf{(b)},
the dropped units are concentrated around zero.
In addition, the other dropped units distribute uniformly, 
i.e., replaceable units are dropped.
Consequently, the large-value units are maintained, i.e., more representative information is maintained.

\subsection{Geometric Pooling}
\label{subsec:gp}
To get rid of the constraints in the analysis of Section \ref{subsec:analysis}, 
we proposed a novel pooling decision criterion - 
Geometric Pooling (GP). GP used the similarity between different node features 
to judge the importance of the node.
The whole pooling process is formulated as follows.

We use the feature concatenation method in DGCNNs\cite{Zhang2018DGCNN} as an example, 
i.e., the input features of a graph are the concatenation of 
the outputs {\color{black}of} different graph convolution layers, 
denoted as $\mathbf{H}^{0:L} \in \mathcal{R}^{n \times d'}$, 
where $n$ is the node number of the graph, 
$d' = \sum_{l = 0}^L d_l$, and $L$ is layer number of the whole network.
$\mathbf{H}^{0:L}$ consists of $\mathbf{H}^{0:L}_j \in \mathcal{R}^{d'} (j = 1, \ldots, n)$, 
which denotes the concatenated feature of each node.
Then, the similarity (distance) between different node features is computed as follows.
\begin{equation}
\label{eq:compute_similarity}
\mathbf{S}_j = \sum_{i = 1}^n \|\mathbf{H}^{0:L}_i - \mathbf{H}^{0:L}_j\|,
\end{equation}
where $\mathbf{S} = \{\mathbf{S}_j\}_{j = 1}^n$ denotes the similarity vector of all nodes, 
and $\|\cdot\|$ denotes the metric used to measure the similarity.

According to $\mathbf{S}$, we sorted the $\{\mathbf{S}_j\}_{j = 1}^n$ 
in Eq. \ref{eq:compute_similarity} to get the node features with least similarity:
\begin{equation}
\label{eq:sort_similarity}
\mathbf{Idx} = Topk(\{\mathbf{S}_j\}_{j = 1}^n).
\end{equation}
In Eq. \ref{eq:sort_similarity}, $Topk$ is an operator to select the index of 
the top-k minimal similarity and the selected indices are saved in $\mathbf{Idx}$.
$k$ is a hyper-parameter, which means the number of maintaining node features,
to make sure the node number of graphs be same.
By $\mathbf{Idx}$, we maintain the node features with the index in $\mathbf{Idx}$.
The whole pooling process is stated in Alg. \ref{alg:gp}.

\begin{algorithm}
	\caption{Geometric Pooling}
	\label{alg:gp}
	\begin{algorithmic}[1]
		\Require k - the number of maintaining nodes
		\Statex \textbf{Input:} $\mathbf{H}^{0:L} = \{\mathbf{H}^{0:L}_j\}_{j = 1}^n$
		\For{$1 \leq j \leq n$}
		\State Compute $\mathbf{S}_j$ by Eq. \ref{eq:compute_similarity}
		\EndFor
		\State Sort $\mathbf{S} = \{\mathbf{S}_j\}_{j = 1}^n$
		\State Compute $\mathbf{Idx}$ by Eq. \ref{eq:sort_similarity}
		\State Select the node feature with index in $\mathbf{Idx}$
		\Statex \textbf{Output:} $\{\mathbf{H}^{0:L}_k\}_{k \in \mathbf{Idx}}$
	\end{algorithmic}
\end{algorithm}

\subsection{Entropy Reduction: distribution drag}
\label{subsec:entropy_reduction}

\begin{figure*}[thb]
	\centering
	\includegraphics[scale=0.62]{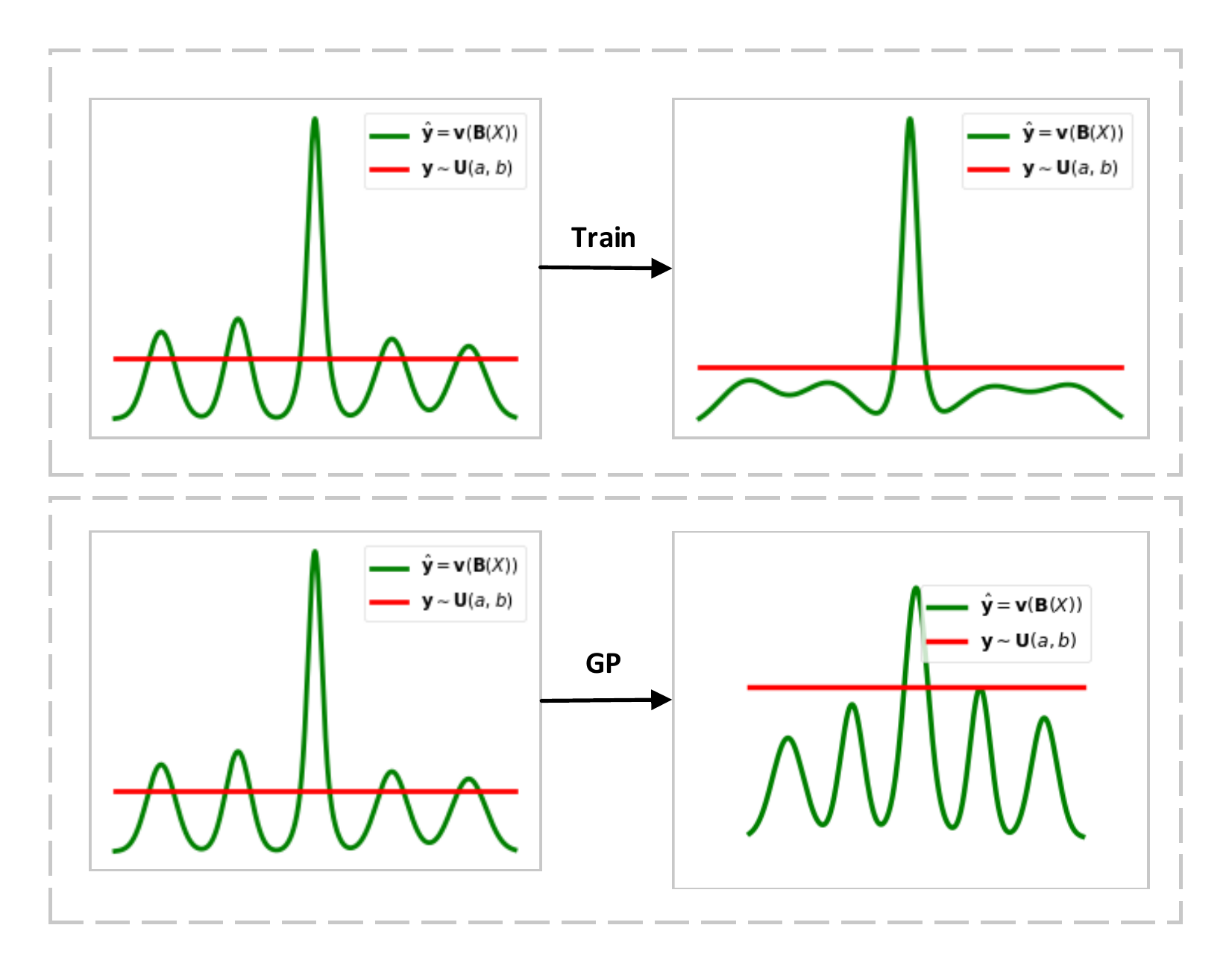}
	\caption{We explain the effectiveness of GP from the output distribution drag view.
		Typically, the common cross entropy loss singly encourages the output distribution
		close to the distribution of the ground truth in a one-hot manner. 
		This results in an over-fitting problem. 
		By GP, the output distribution is encouraged to be close to the uniform distribution, 
		i.e., it is punished to be far away from the ground truth.
    {\color{black}
    $\hat{y}=v(\mathbf{B}(x))$ represents the output of GNNs, where $\mathbf{B}$ denotes 
    the feature extractor of GNNs and $v$ denotes the final linear layer. 
    $y \sim \mathbf{U}(a, b)$ denotes samples over a uniform distribution between $(a, b)$. }
    }
	\label{fig:distribution_drag}
\end{figure*}

The typical convolutional neural networks gradually strengthen 
the influence of the training samples, i.e., the distribution of output logits 
is forced far away from the initial uniform distribution.
We visualized this process in Figure \ref{fig:distribution_drag}.

To state the problem, we first give the formulation of the typical 
loss function - Cross Entropy Loss:
\begin{equation}
\label{eq:cross_entropy_loss}
\mathcal{L}_{CE} = \sum_{\mathbf{x}_i} \mathbf{y}_i \log \frac{1}{\mathbf{q}(\mathbf{x}_i)} = -\sum_{\mathbf{x}_i} \mathbf{y}_i \log \mathbf{q}(\mathbf{x}_i),
\end{equation}
where $\mathbf{x}_i$ is the training sample; 
{\color{black}$\mathbf{q}(\mathbf{x}_i)$} denotes the output of the model;
$\mathbf{y}_i$ is the label of $\mathbf{x}_i$.
When $\mathcal{L}_{CE}$ decreases, the distribution of $\mathbf{y}_i$ is 
pulled to be close to {\color{black}$\mathbf{q}(\mathbf{x}_i)$}, which is in a one-hot manner.
The one-hot label is an extreme probability formulation that is far away from the 
uniform distribution. 

Although the one-hot label is the target of the task, 
a lot of regularization research shows that 
just pulling the distribution close to the one-hot distribution  
results in an over-fitting problem. 
For example, Miller \cite{Mller2019WhenDL} proposed 
that the effectiveness of Label Smoothing results from the distribution drag.
Particularly, the one-hot label is modified to a probability form: 
$$[0, 0, 1, 0, 0] \rightarrow [0.05, 0.05, 0.8, 0.05, 0.05].$$
Obviously, $\mathbf{q}(\mathbf{x}_i)$ is enforced to be close to the uniform distribution.
Label Smoothing effectively strengthens the robustness 
by alleviating the too-high confidence problem
at the training stage, i.e., the over-fitting problem \cite{2019CVPRReLU}.

Many similar units filtered by the convolutional kernels jointly output
the final decision of the too-high confidence \cite{2019CVPRFilterPruning}.
Hence dropping the units of high similarity alleviates
the too-high confidence problem.
Except for the pooling operator, the regularization effectiveness of 
the proposed GP can be viewed as a penalty term:
\begin{equation}
\begin{split}
\mathbf{L} & = \mathbf{L}_{CE} + \mathbf{L}_{ana} \\
&= -\sum_{\mathbf{x}_i} \left[\mathbf{y}_i \log \mathbf{q}(\mathbf{x}_i) + \lambda \cdot D_{KL}(\mathbf{U}_{[0, C]} \| \mathbf{q}(\mathbf{x}_i)) \right]
\end{split}
\end{equation}
where $D_{KL}(\cdot\|\cdot)$ is a Kullback-Leibler divergence; $\lambda$ is the scaling 
factor; $C$ is the class number of the task. 
We give the $D_{KL}(\mathbf{U}_{[0, C]} \| \mathbf{q}(\mathbf{x}_i))$ as follows.
\begin{equation}
\begin{split}
D_{KL}(\mathbf{U}_{[0, C]} \| \mathbf{q}(\mathbf{x}_i)) &= -\mathbf{U}_{[0, C]}\log \frac{\mathbf{q}(\mathbf{x}_i)}{\mathbf{U}_{[0, C]}} \\
&= \frac{1}{C} \log \frac{1}{C} - \frac{1}{C} \log \mathbf{q}(\mathbf{x}_i)
\end{split},
\end{equation}
where $\frac{1}{C} \log \frac{1}{C}$ is the negative entropy of 
the uniform distribution $\mathbf{U}_{[0, C]}$
and $-\frac{1}{C} \log \mathbf{q}(\mathbf{x}_i)$ is the cross entropy between 
$\mathbf{U}_{[0, C]}$ and $\mathbf{q}(\mathbf{x}_i)$.
$\frac{1}{C} \log \frac{1}{C}$ is a constant in a particular task.
When $-\frac{1}{C} \log \mathbf{q}(\mathbf{x}_i)$ decreases, 
the output is enforced to be close to $\mathbf{U}_{[0, C]}$,
and hence the over-fitting problem is alleviated.
To clearly show the distribution drag, we visualized the distribution 
variation of the vanilla training in the above row and 
the proposed GP in the below row.

\section{Experiments}
To show the effectiveness of the proposed GP, we embedded GP in 
different methods to replace the original pooling technologies on the typical 
graph classification dataset - TUdataset.

\subsection{Set Up}
\label{subsec:exp_set_up}
\textbf{Datasets.}
TUdataset consists of too many sub-datasets. To ensure a fair comparison, {\color{black}we follow the settings of the previous method DGCNNs \cite{Zhang2018DGCNN}.}
Hence we chose MUTAG, PTC, NCI1, PROTEINS, D\&D, COLLAB, IMDB-B, 
and IMDB-M to evaluate the performance.
Most chosen sub-datasets contain graphs of two categories, except for COLLAB and 
IMDB-M, which contain graphs of three categories.
The detailed information of the sub-datasets is given in 
Table \ref{tab:dataset_information}.
%\begin{table}[h]
%    \centering
%    \small
%    \setlength{\tabcolsep}{1mm}{
%    \caption{The dataset description.}
%    \begin{tabular}{l|cccccccc}
%    \toprule[1.2pt]
%        Dataset & MUTAG & PTC & NCI1 & PROTEINS & D\&D & COLLAB & IMDB-B & IMDB-M \\
%    \hline
%        \#Graphs & 188 & 344 & 4,110 & 1,113 & 1,178 & 5,000 & 1,000 & 1,500\\
%        \#Classes & 2 & 2 & 2 & 2 & 2 & 3 & 2 & 3\\
%        \#Node Attr. & 8 & 19 & 38 & 5 & 90 & 1 & 1 & 1 \\
%    \hline
%    \end{tabular}}
%    \label{tab:dataset_information}
%\end{table}

\begin{table}[h]
	\centering
	\small
	\caption{The dataset description.}
	\begin{tabular}{l|cccc}
		\toprule[1.2pt]
		Dataset & MUTAG & PTC & NCI1 & PROTEINS \\ 
		\hline
		\#Graphs & 188 & 344 & 4,110 & 1,113 \\
		\#Classes & 2 & 2 & 2 & 2 \\
		\#Node Attr. & 8 & 19 & 38 & 5 \\
		\hline
		Dataset & D\&D & COLLAB & IMDB-B & IMDB-M \\
		\#Graphs & 1,178 & 5,000 & 1,000 & 1,500\\
		\#Classes & 2 & 3 & 2 & 3\\
		\#Node Attr. & 90 & 1 & 1 & 1 \\
        \toprule[1.2pt]
	\end{tabular}
	\label{tab:dataset_information}
\end{table}

\textbf{Comparison methods.}
We compare the proposed GP with sort pooling in DGCNNs \cite{Zhang2018DGCNN}, 
SAGPooling \cite{Lee2019SAGPooling}, DiffPooling \cite{Ying2018DiffPooling}, 
and GRAPHSAGE \cite{Hamilton2017GRAPHSAGE} to show GP's effectiveness.
Also, several kernel-based algorithms are compared, including GRAPHLET
\cite{Shervashidze2009GRAPHLET}, Shortest-Path \cite{Borgwardt2005SHORTESTPATH},
Weisfeiler-Lehman kernel (WL) \cite{Shervashidze2011WL}, 
and Weisfeiler-Lehman Optimal Assignment Kernel (WL-OA) \cite{Kriege2016WLOA}.

\textbf{Model Configuration.}
We used five Graph Convolutional layers to extract different-level
features of the graph and a GP layer is used to obtain the salient features and 
make each graph share the same node number and the same node feature number.
The hyper-parameter $k$ in Eq. \ref{eq:sort_similarity} is varying for different datasets.
We set the $k$ such that 60\% graphs have nodes more than $k$.
For fair evaluation, we used a single network structure on all
datasets and ran GP using exactly the same folds as
used in graph kernels in all 100 runs of each dataset.

\subsection{Comparison with deep approaches}
As given in Table \ref{tab:results_deep}, the results of the compared 
deep approaches show the effectiveness of the proposed GP and its 
variation -- GP-mixed.
GP-mixed first used sort pooling \cite{Zhang2018DGCNN} to drop some units 
contributing little to the final decision.
Then GP is used to drop the replaceable units.

\begin{table}[htb]
	\centering
	\small
	\renewcommand{\arraystretch}{1.2}
	\caption{The performance comparison among different deep approaches.
		The top two test classification accuracys on each dataset are in bold.}
	\begin{tabular}{lcccc}
		\toprule[1.2pt]
		\multirow{2}{*}{\textbf{Methods}} & \multicolumn{4}{c}{\textbf{Datasets}} \\
		\cline{2-5}
		& MUTAG & PTC & NCI1 & PROTEINS  \\
		\hline
		DGCNNs & 85.83 & 58.59 & 74.44 & 75.54  \\
		GRAPHSAGE & -- & -- & -- & 70.48 \\
		SAGPooling & -- & -- & 74.06 & 70.04 \\ 
		DiffPooing & -- & -- & -- & 76.25  \\
		\hline
		GP & \textbf{86.10} & \textbf{61.30} & \textbf{75.80} & \textbf{77.10} \\
		GP-mixed & \textbf{86.44} & \textbf{61.86} & \textbf{76.29} & \textbf{77.64} \\
		\toprule[1.2pt]
		& D\&D & COLLAB & IMDB-B & IMDB-M \\
		DGCNNs & 79.37 & 73.76 & 71.50 & 46.47 \\
		GRAPHSAGE & 75.42 & 68.25 & 68.80 & 47.6 \\
		SAGPooling & 76.19 & 69.47 & -- & -- \\
		DiffPooing & 79.64 & \textbf{75.48} & -- & -- \\
		\hline
		GP & \textbf{80.21} & 74.63 & \textbf{72.20} & \textbf{47.89} \\
		GP-mixed & \textbf{80.28} & \textbf{75.50} & \textbf{72.68} & \textbf{48.32} \\
		\toprule[1.2pt]
	\end{tabular}
	\label{tab:results_deep}
\end{table}

In Table \ref{tab:results_deep}, 
we give several state-of-the-art GNNs pooling performance.
For example, on D\&D, the proposed GP outperforms DGCNNs and SAGPooling by 0.84\% and 4.02\%, 
respectively. On COLLAB, GP outperforms DGCNNs by 0.87\%. 
Although GP underperforms DiffPooling, GP-mixed slightly outperforms DiffPooling.
In general, the proposed GP outperforms the SOTA deep approaches by 0.7\% $\sim$ 5.05\%.

\subsection{Comparison with kernel-based approaches}
Kernel-based approaches are also compared.
As shown in Table \ref{tab:results_kernels}, 
the proposed GP and its variant outperform the kernel-based 
approaches on most evaluation datasets.
For example, GP-mixed outperforms the compared kernel-based approaches by 
1.21\% $\sim$ 4.73\% on PROTEINS.
Unfortunately, on COLLAB, GP and GP-mixed underperform WL and WL-OA.
\begin{table*}[htb]
	\centering
	\small
	\renewcommand{\arraystretch}{1.2}
	\caption{The performance comparison between GP and kernel-based approaches.
		The top two test classification accuracies on each dataset are in bold.}
	\begin{tabular}{lcccc}
		\toprule[1.2pt]
		\multirow{2}{*}{\textbf{Methods}} & \multicolumn{4}{c}{\textbf{Datasets}} \\
		\cline{2-5}
		& MUTAG & PTC & NCI1 & PROTEINS \\
		\hline
		GRAPHLET & 85.7 & 54.7 & \textbf{76.20} & 72.91 \\
		SHORTEST-PATH & 83.15 & -- & -- & 76.43 \\
		WL & 84.11 & 57.97 & 84.46 & 73.76 \\
		WL-OA & 84.50 & -- & -- & 75.26 \\
		\hline
		GP & \textbf{86.10} & \textbf{61.30} & 75.80 & \textbf{77.10} \\
		GP-mixed & \textbf{86.44} & \textbf{61.86} & \textbf{76.29} & \textbf{77.64} \\
		\toprule[1.2pt]
		& D\&D & COLLAB & IMDB-B & IMDB-M \\
		GRAPHLET & 74.85 & 64.66 & -- & -- \\
		SHORTEST-PATH & 78.86 & 59.10 & -- & -- \\
		WL & 74.02 & \textbf{78.61} & 71.26 & 46.52\\
		WL-OA & 79.04 & \textbf{80.74} & -- & -- \\
		\hline
		GP & \textbf{80.21} & 74.63 & \textbf{72.20} & \textbf{47.89} \\
		GP-mixed & \textbf{80.28} & 75.50 & \textbf{72.68} & \textbf{48.32} \\
		\toprule[1.2pt]
	\end{tabular}
	\label{tab:results_kernels}
\end{table*}

\subsection{Comparison on the computation cost}
To show the simplification of the proposed GP, we also compute the 
computation cost of the compared deep approaches.
\begin{figure}
	\centering
	\includegraphics[scale=0.56]{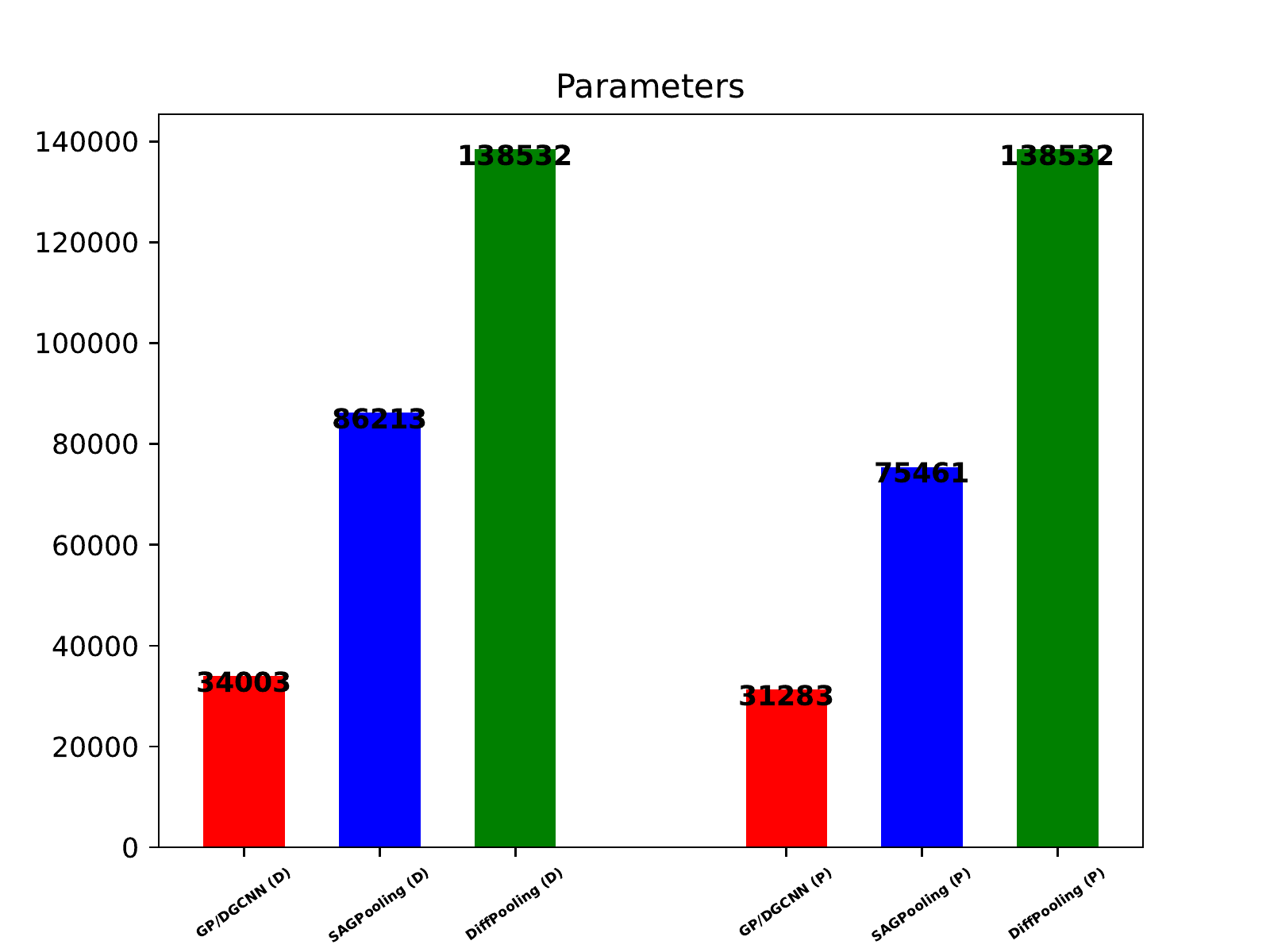}
	\caption{Parameter comparison. We give the parameters of different deep approaches on
		two datasets: D\&D and PROTEINS. For example, let \textbf{GP/DGCNNs (D)} denote 
		the parameter amount of the proposed GP and DGCNNs on the D\&D dataset. 
		Also, \textbf{SAGPooling (P)} denotes the parameter amount of SAGPooling
		on PROTEINS dataset. 
		The numbers on the colored bars mean the exact parameter number.}
	\label{fig:params_comparison}
\end{figure}
As shown in Figure \ref{fig:params_comparison}, the red bars, which mean the 
parameter amount of the proposed GP, are far less than that of SAGPooling and 
DiffPooling approaches. On D\&D, the parameter amount of GP accounts for 39.44\% of 
SAGPooling and 24.55\% of DiffPooling.
This shows the computation efficiency of the proposed GP.

\section{Limits and Conclusion}
{\color{black}
    The proposed GP assumes that different nodes in a graph do not share too high similarity, 
    i.e., the similar nodes are classified in the same category.    
    Thus, GP may not be suitable for a strongly connected cluster in a graph, 
    whose nodes in this cluster share a high similarity. 
}

In the work, we proposed a novel graph pooling technology based on the 
geometric distance between different node features, called Geometric Pooling (GP).
The proposed GP maintains more representative information, which is dropped by 
other global graph pooling technologies.
By combining the sorting pooling and GP, we proposed a more effective pooling method, called GP-mixed, bringing a higher classiﬁcation accuracy.
In addition, the replaceable feature drop reduces the entropy reduction, 
which encourages the output distribution away from the hard targets.
This alleviates the over-confidence problem and hence 
address the over-fitting problem.

%%%%%%%%%%%%%%%%%%%%%%%%%%%%%%%%%%%%%%%%%%
\bibliographystyle{plain}
\bibliography{GPooling}

\clearpage

\appendix
\section{Activation Function Replacement}
\label{appendix:sec:activation}

We replaced the activation function \textit{Tanh} in DGCNNs with 
ReLU (DGCNNs-ReLU in table) and observed a performance drop in Table \ref{tab:results_activation}.
The results showed that DGCNNs-ReLU underperforms DGCNNs by 1.2\%.

\begin{table}[htb]
	\centering
	\small
	\renewcommand{\arraystretch}{1.2}
	\caption{\textit{Tanh v.s. ReLU}}
	\begin{tabular}{lccccc}
		\toprule[1.2pt]
		\multirow{2}{*}{\textbf{Methods}} & \multicolumn{4}{c}{\textbf{Datasets}} & Avg. $\downarrow$ \\
		\cline{2-6}
		& MUTAG & PTC & NCI1 & PROTEINS  & \multirow{3}{*}{1.24}\\
		\cline{1-5}
		DGCNNs & 85.83 & 58.59 & 74.44 & 75.54  \\
		\cline{1-5}
		DGCNNs-ReLU & 83.24 & 58.32 & 73.48 & 74.40 \\
		\toprule[1.2pt]
		& D\&D & COLLAB & IMDB-B & IMDB-M & \multirow{3}{*}{1.27}\\
        \cline{1-5}
		DGCNNs & 79.37 & 73.76 & 71.50 & 46.47 \\
		\cline{1-5}
		DGCNNs-ReLU & 77.74 & 71.56 & 71.32 & 45.40 \\
		\toprule[1.2pt]
	\end{tabular}
	\label{tab:results_activation}
\end{table}

\section{Ablation Study}

\textbf{Different similarity metrics.}
We study the influence of different metrics on 
the similarity of different node features.
As given in Table \ref{tab:results_metrics}, 
Euclidean distance gets the best test accuracy.
As argued in TSMA \cite{Wang2022TSMA}, 
the scalar item of Euclidean distance 
(i.e. $||x||^2 + ||y||^2$ in $||x, y||^2 = ||x||^2 + ||y||^2 - 2\cdot(x\cdot y)$)
is helpful to accelerate the model training.
We think it is also helpful to measure the similarity of 
different nodes.

\begin{table}[htb]
	\centering
	\small
	\renewcommand{\arraystretch}{1.2}
	\caption{We used Euclidean distance, 
    inner product, and cosine similarity in GP for ablation.
    $\mathcal{L}_2$, $IP$ (Inner Product), and $Cos.$ are used 
    to represent the above metrics respectively.}
	\begin{tabular}{lcccc}
		\toprule[1.2pt]
		\multirow{2}{*}{\textbf{Methods}} & \multicolumn{4}{c}{\textbf{Datasets}}  \\
		\cline{2-5}
		& MUTAG & PTC & NCI1 & PROTEINS  \\
		\cline{1-5}
		$\mathcal{L}_2$ & 86.10 & 61.30 & 75.80 & 77.10  \\
		\cline{1-5}
		$IP$ & 86.04 & 60.85 & 75.42 & 76.68 \\
        \cline{1-5}
		$Cos.$ & 85.70 & 59.06 & 75.04 & 76.05 \\
		\toprule[1.2pt]
		& D\&D & COLLAB & IMDB-B & IMDB-M \\
        \cline{1-5}
		$\mathcal{L}_2$ & 80.21 & 74.63 & 72.20 & 47.89 \\
		\cline{1-5}
		$IP$ & 80.06 & 74.88 & 72.00 & 47.47 \\
		\cline{1-5}
		$Cos.$ & 79.64 & 74.76 & 71.76 & 46.83 \\
		\toprule[1.2pt]
	\end{tabular}
	\label{tab:results_metrics}
\end{table}

\EOD

\end{document}